\documentclass[10pt,twocolumn,letterpaper]{article}

\usepackage{cvpr}              % To produce the CAMERA-READY version
\usepackage{tabularx}
\usepackage{siunitx} 
\usepackage{graphicx}   % Required for \resizebox
\usepackage{caption} % Required for \captionof
\usepackage{xcolor,colortbl}
\usepackage{tcolorbox}
\usepackage{enumitem}
\usepackage{amssymb}

\definecolor{first_color}{HTML}{FFDFBF}
\definecolor{second_color}{HTML}{FFFBCC}
\definecolor{third_color}{HTML}{FFDFBF}
\newcommand{\first}[1]{{\cellcolor{first_color} #1}}
\newcommand{\second}[1]{{\cellcolor{second_color} #1}}

\usepackage{xcolor}
\usepackage{amssymb}

\newcommand{\cmark}{\textcolor{green!60!black}{$\checkmark$}}
\newcommand{\xmark}{\textcolor{red!60!black}{$\times$}}

\definecolor{cvprblue}{rgb}{0.21,0.49,0.74}
\usepackage[pagebackref,breaklinks,colorlinks,allcolors=cvprblue]{hyperref}

\def\paperID{*****} % *** Enter the Paper ID here
\def\confName{3DV\xspace}
\def\confYear{2027\xspace}

\title{Gen2Physics: Grounding Generated 3D Meshes in Physics via Multi-View Material Decomposition}

\author{
Mauro Comi$^{1, 3}$ \and Jordi Serrano Berbel$^{1}$ \and Kevis-Kokitsi Maninis$^{1}$ \and Philipp Henzler$^{2}$ \and Manuel Sanchez$^{1}$ \\ \\ $^{1}$Google DeepMind \ \ $^{2}$Google Research \ \  $^{3}$University of Bristol \\
}

\begin{document}

\twocolumn[{%
\renewcommand\twocolumn[1][]{#1}%
\maketitle
\centering
\includegraphics[width=0.85\linewidth]{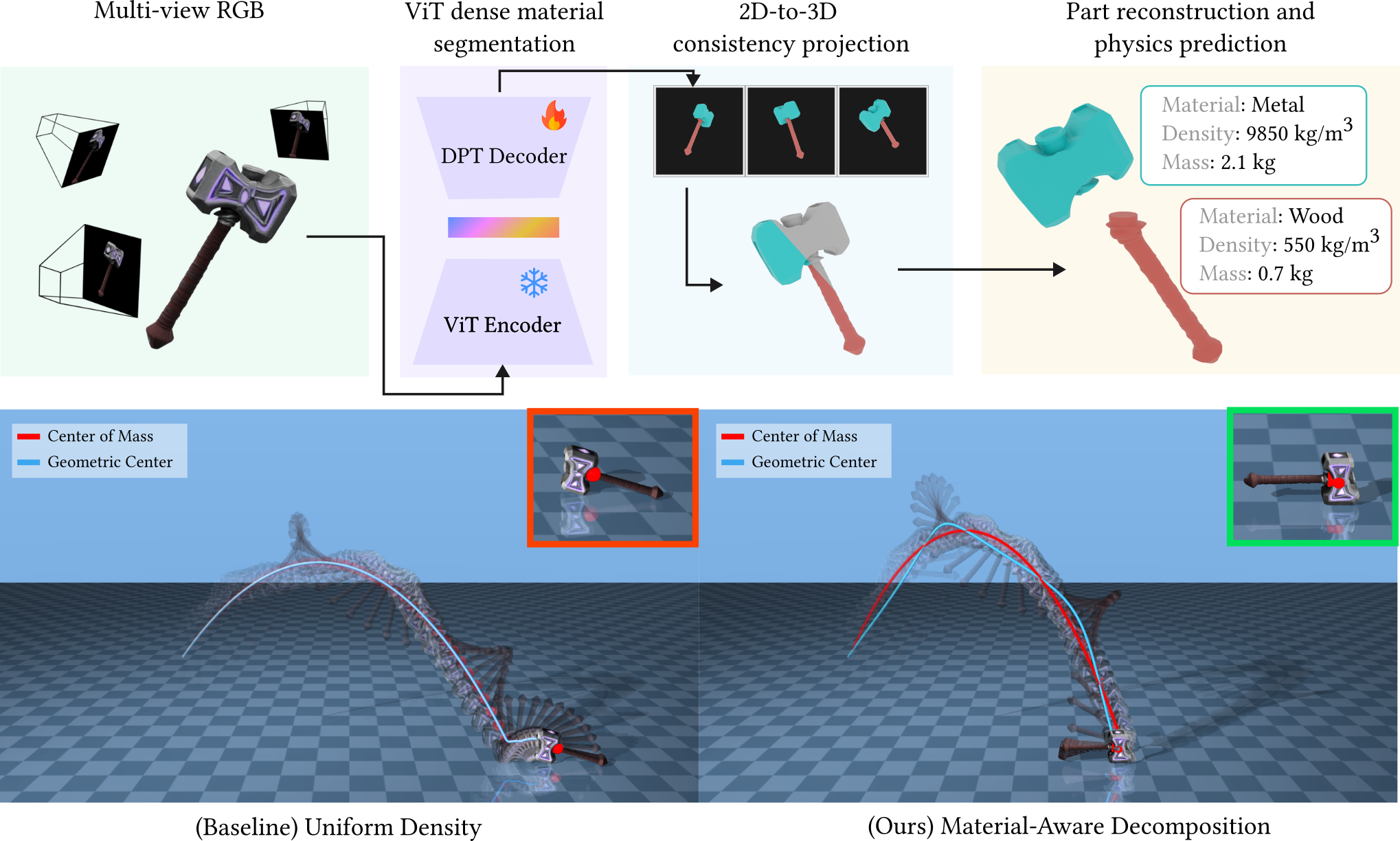}
\captionof{figure}{Our framework decomposes a static mesh into simulation-ready components. Top: Our pipeline uses a Dense Prediction Transformer (DPT) to predict material labels from multi-view 2D views, and then it aggregates the predictions into a consistent 3D material map. This guides a 3D part reconstruction model to generate watertight components and assigns distinct physics properties to each part. Bottom: A comparison of dynamics in a free-fall MuJoCo simulation. The baseline (left) assumes uniform density and produces physically inaccurate rotation around the geometric center. Using an asset produced by our method (right), MuJoCo correctly identifies the Center of Mass within the heavy metal head and results in more accurate dynamics.}
\label{fig:figure1}
}
\vspace{1em}
]

\begin{abstract}
While state-of-the-art generative models produce high-fidelity 3D meshes, these outputs lack the physical properties required for interactive simulation, gaming, or robotics. We introduce Gen2Physics, a unified and automated framework that grounds generated meshes in physics by automatically decomposing them into their constituent material components. Unlike prior approaches, which focus on volumetric representations incompatible with standard physics engines, Gen2Physics operates directly on meshes to produce immediately simulation-ready assets. Our pipeline integrates a fine-tuned Vision Transformer for dense material segmentation, a robust 2D-to-3D consistency projection, and a Vision-Language Model (VLM) guided refinement that leverages contextual reasoning to assign physical properties and infer internal geometry (solid vs. hollow). By converting surface patches into volumes with distinct densities, our method enables physically plausible dynamic simulations. Experimental results on the ABO-500 and PartNet-Material benchmarks demonstrate that Gen2Physics more than doubles the material segmentation accuracy of prior physics-grounding pipelines (15.6 to 48.3 mIoU), while matching the mass-estimation accuracy of volumetric methods and being the only approach to output watertight per-material sub-meshes.
\end{abstract}
\section{Introduction}
\label{sec:intro}
State-of-the-art 3D generative models like TRELLIS \cite{xiang2025structured} or SPARC3D \cite{li2025sparc3d} can produce visually stunning meshes from single images \cite{hong2023lrm}, yet their outputs are static assets, unsuitable for interactive applications. This results from a lack of physical properties necessary for realistic interactive behavior. An important gap exists between visual fidelity and physical plausibility, limiting their use in robotics, gaming, and simulation.

Existing approaches \cite{zhai2024physical, shuai2025pugs} address this gap through physics-aware property estimation techniques. These methods are powerful but operate on volumetric representations such as Neural Radiance Fields (NeRFs \cite{mildenhall2021nerf}) or 3D Gaussian Splatting (3DGS \cite{kerbl20233d}). While suitable for particle or continuum simulators, these representations are incompatible with standard rigid-body and articulated multi-body physics engines used in robotics, gaming, and simulation (e.g., MuJoCo \cite{todorov2012mujoco}, NVIDIA Isaac Sim, Bullet \cite{coumans2010bullet}), which expect discrete surface or convex sub-meshes with distinct inertial and contact properties. While it is possible to extract a mesh from these formats, the process produces a monolithic mesh. For example, a multi-material object (e.g. a wooden chair with metal legs) would be extracted as one continuous surface, making it difficult to automatically assign distinct physical properties to its material components. Although recent part segmentation frameworks \cite{lin2025partcrafter, liu2025partfield} address object decomposition, they focus on semantic categorization (e.g. "a mug", "a handle") rather than material composition. This distinction is important, as an object's physical behavior is governed by its material properties rather than uniquely its categorical label. Therefore, while powerful individual components exist in the literature, it remains an open challenge to take an existing mesh and recover the material partition a rigid-body simulator needs, which may not align with its semantic parts.

To address these limitations, we introduce \textit{Gen2Physics}, a unified framework that grounds arbitrary generated meshes in physics by automatically producing material-based components usable in standard physics engines. Unlike physics-aware generative approaches~\cite{cao2026physx, cao2026physx3d}, which infer physical properties jointly with geometry, Gen2Physics operates post-hoc and can be applied to assets from any generator without retraining. Our core hypothesis is that an object’s visual material and its global context are a strong proxy for its physical properties. Gen2Physics fine-tunes a state-of-the-art 2D Vision Transformer \cite{dosovitskiy2020image} (TIPS~\cite{maninis2024tips}) for dense material segmentation on a large corpus of annotated material segmentation masks. Inspired by prior work \cite{mccormac2017semanticfusion}, we propose a 2D-to-3D consistency projection method that robustly lifts 2D material segmentations into the 3D domain, resolving ambiguities present in any single view. This allows us to automatically decompose a 3D mesh into its constituent material components, complete them into watertight volumes, and assign physical properties compatible with physics engines. We demonstrate that Gen2Physics enables plausible physics simulations for assets produced by state-of-the-art 3D generative models. In this work we specifically target rigid-body dynamic simulation governed by per-part mass distribution, center of mass, inertia tensors, rather than deformable continuum elasticity which relies on volumetric parameters like Young's modulus and Poisson's ratio.

Our contributions can be summarized as follows:
\begin{itemize}[topsep=0pt]
\item We introduce Gen2Physics, a unified pipeline that grounds existing 3D meshes in physics by automatically decomposing them into simulation-ready components.
\item We demonstrate a robust methodology for lifting 2D ViT segmentations into a coherent 3D material map using multi-view aggregation and VLM refinement, solving ambiguity issues inherent in dense ViT predictions.
\item We outperform prior baselines in 3D material segmentation and match the mass-estimation accuracy of volumetric methods, while producing per-material sub-meshes.
% \item We introduce the PartNet-Material benchmark, a \textit{manually curated} test set of 1100 multi-view images across 100 objects sampled from the PartNet-Mobility dataset \cite{mo2019partnet}.
\end{itemize}

\section{Related Work}
\subsection{3D Generative Models}
% TRELLIS, SPARC3D, PartCrafter produce high-quality textured meshes in seconds. Gen2Physics is designed to work directly with the output of these models.
Modern 3D generative models are capable of generating high-quality textured meshes from various inputs, including single images. TRELLIS~\cite{xiang2025structured} develops a unified structured latent representation that allows decoding to different output formats, including meshes, from text or image prompts. SPARC3D~\cite{li2025sparc3d} focuses on a sparse deformable Marching Cubes \cite{lorensen1998marching} representation (Sparcubes) and a sparse convolutional VAE (Sparconv-VAE) for high-resolution 3D shape modeling. Gen2Physics is specifically designed to work directly with the output of these models, transforming their static outputs into physically-grounded assets. Moreover, novel methods for reconstructing sub-meshes from 3D segmented meshes have been proposed. HoloPart \cite{yang2025holopart} is currently the state-of-the-art in terms of accuracy and speed.

\subsection{3D Part Segmentation}
Although the field of segmentation has been associated with image masking, models that act on 3D data structures have recently been proposed. SAMPart3D~\cite{yang2024sampart3d}, PartField~\cite{liu2025partfield}, PartCrafter~\cite{lin2025partcrafter} and PartFormer~\cite{tan2024partformer} decompose shapes directly in 3D, targeting semantic and instance parts. However, physical behaviour is dictated by material composition rather than semantic identity. For example, "chair leg" may span multiple materials, so a semantic partition underdetermines the parameters a simulator needs. These methods are also clustering-based rather than identification-based, taking the number of parts as input a-priori~\cite{liu2025partfield}, whereas the number of distinct materials is often unknown and must be discovered by the model.

Pairing an open-world segmenter such as SAM~\cite{kirillov2023segment} with a VLM classifier is a natural alternative, but such segmenters follow object and topological boundaries, which may not coincide with material transitions. In contrast, our fine-tuned ViT instead predicts dense per-pixel material classes across geometry-agnostic boundaries.

A related line of work (MaterialSeg3D~\cite{li2024materialseg3d}) fuses multi-view predictions from a dense 2D material segmenter into a 3D material assignment, targeting \textit{optical} properties. In contrast, we predict \textit{mechanical} properties, assigning per-face labels that drive part decomposition, amodal completion and per-part mass. Moreover, fusing on faces rather than in UV space removes any dependence on an existing UV parameterisation.

\subsection{Physics from 3D Segmentation}

A growing body of literature aims to integrate physics into 3D models using Vision Language Models (VLMs) to reason about object properties. NeRF2Physics~\cite{zhai2024physical} extracts 3D points from a neural radiance field and fuses 2D vision-language features to estimate physical properties, including mass density. PUGS~\cite{shuai2025pugs} extends this concept to 3DGS representations, aiming for more efficient reconstruction and physical property inference. PhysGen3D~\cite{chen2025physgen3d} transforms a single image into an interactive 3D scene by estimating 3D shapes, poses, and physical properties, using Material Point Methods (MPM) \cite{sulsky1995application} for simulation and GPT-4o for property estimation. Phys4DGen~\cite{lin2024phys4dgen} also integrates physics simulation directly into the 4D generation pipeline, notably aiming to perceive multi-material compositions and internal structures. Concurrent work \cite{dagli2025vomp} regresses continuous volumetric mechanical property fields for deformable simulation, and Pixie \cite{le2025pixie} trains a feed-forward network that predicts a voxel-grid material field for MPM simulation.

These approaches commonly rely on volumetric representations, such as NeRFs, Gaussians Splatting, per-voxel features, or particle-based MPM, for physical property estimation. While volumetric representations can be coupled with continuum or particle-based simulators (e.g., MPM \cite{sulsky1993particle} or FEM), standard rigid-body and articulated multi-body engines (e.g., MuJoCo \cite{todorov2012mujoco}, Bullet \cite{coumans2010bullet}) expect sub-meshes with explicit inertial and contact boundaries. Converting continuous volumetric density fields into distinct, multi-material mesh partitions remains an ill-posed and error-prone challenge \cite{li2025sparc3d, lorensen1998marching}. In contrast, Gen2Physics directly operates on generated meshes, making its output inherently compatible with standard physics engines.
\section{Method}
Our goal is to decompose a 3D mesh $M$, such as one produced by a generative model like TRELLIS, into a set of physically-grounded and simulation-ready components $\{P_1, P_2, ..., P_k\}$. 
Each $P_i$ is a watertight sub-mesh corresponding to a distinct material, annotated with plausible physical properties. Our method, Gen2Physics, achieves this through a four-stage pipeline:  (1) dense material segmentation from multiple views using a specialized Vision Transformer; (2) robust 2D-to-3D projection to establish a geometrically consistent material labeling of the mesh; (3) semantic refinement of these labels using a Vision-Language Model; and (4) amodal completion (i.e. inferring the complete shape including occluded or internal geometry) of material parts and assignment of physical properties.

\subsection{Multi-View Material Segmentation}
\label{subsec:multi-view}
% - Vision Transformer (ViT) fine-tuned for dense material segmentation.
% - At test time, given a new mesh $M$, we place it in a virtual scene and render it from $N$ viewpoints.
% - We apply our fine-tuned ViT to produce a 2D material segmentation map $S_i$ for each rendered image $I_i$. This might contain view-dependent ambiguities, lighting effects, or model errors
Our first step leverages the rich semantic knowledge encoded in 2D Vision Transformers to infer material properties from RGB renders. To this end, we built a large-scale, object-centric dataset before fine-tuning a state-of-the-art model (TIPS \cite{maninis2024tips}).

\subsubsection{Data Collection for Material Segmentation}
\label{subsubsec:data-collection}
Material segmentation suffers from a scarcity of large-scale, object-centric datasets with detailed material annotations suitable for physics-aware 3D reconstruction. To address this, we curated a dataset of 500,000 multi-view renderings by sourcing assets from Coohom \cite{Coohom3DModels}, a large proprietary repository of 3D assets for interior design. Unlike Objaverse-XL \cite{deitke2023objaverse}, which contains raw unannotated scans, Coohom is composed of assets pre-decomposed into semantically meaningful components, providing a strong foundation for part-based material labeling. 

Our initial procedure for annotating material information consisted of extracting the material shader parameters (i.e. albedo, metallic, specular values) to infer a suitable material label from a list of closed categories. However, our attempt proved unreliable as shader parameters are often tuned for aesthetic appeal rather than physical accuracy. Therefore, we developed an automated pipeline to re-label 20,000 assets with physically-grounded material categories. This set of material categories was adapted from the PACO dataset~\cite{ramanathan2023paco}, a benchmark for part-level understanding. We curated the list of materials proposed by PACO for physical realism by removing visually ambiguous classes (e.g. \textit{Rattan}) and adding functionally distinct ones (e.g. \textit{Rubber}) to better suit interactive scenarios. The full list of materials is [\textit{Wood}, \textit{Rubber}, \textit{Fabric}, \textit{Organic}, \textit{Leather}, \textit{Metal}, \textit{Paper}, \textit{Plastic}, \textit{Glass}, \textit{Ceramic}, \textit{Stone}, \textit{Other}]. The material labeling procedure is as follows:
\begin{enumerate}
    \item For each multi-part asset, we render part-level identity segmentation masks using the high-fidelity renderer Unreal Engine \cite{unrealengine}
    \item Each isolated part is rendered from multiple viewpoints, which were randomly sampled from a spherical distribution around the asset to ensure comprehensive visual coverage.
    \item We leverage a multimodal VLM (Gemini 2.5 Flash \cite{comanici2025gemini}) to assign one of 12 predefined material categories to each part on its multi-view renderings. 
\end{enumerate}
This process resulted in a dataset of 500,000 multi-view renderings with dense material annotation, which allowed us to acquire the rich material understanding necessary for dense material segmentation.

\begin{figure*}[t!]
\centering{\includegraphics[width=1\textwidth]{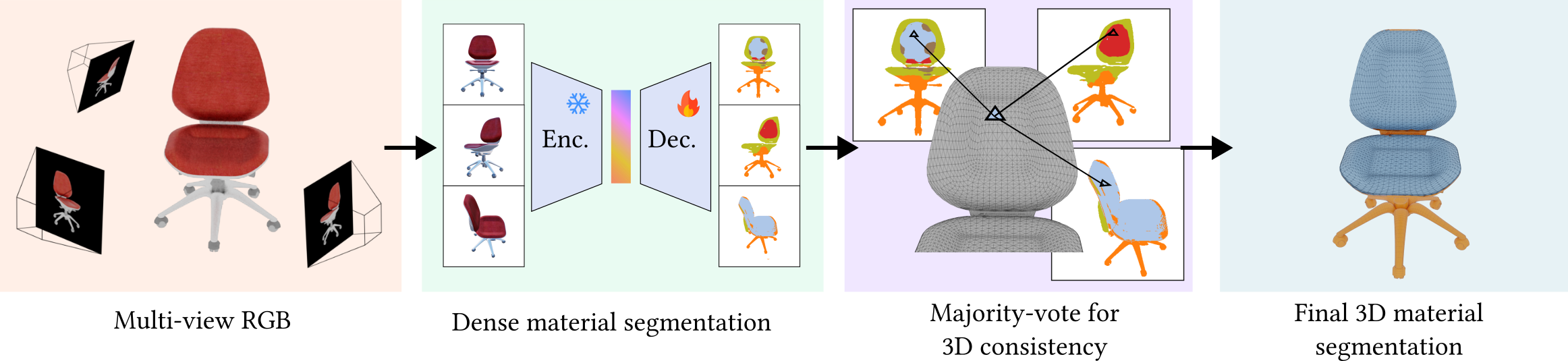}}
  \caption{Process of our 2D-to-3D consistency projection. A fine-tuned ViT predicts material segmentation masks, which are possibly noisy and not consistent across views. We project each face of our asset onto the rendered views, taking into account occlusions. For each face, we assign a label corresponding to the material with the highest frequency in the segmentation masks. The final material prediction is consistent across views. We then re-render the segmentation masks. }
  % \Description{Four panels: a chair being rendered from 3 cameras. Second panel: the renderings are segmented, but the segmentations are not correct. Third pane, a triangle is projected to the three objects. Fourth shows clean segmentation in 3D.}
  \label{fig:projection}
\end{figure*}

\subsubsection{Fine-tuning for Dense Material Segmentation}
% The dataset collected in the previous step was used to train the decoder of the Vision Transformer TIPS for dense material segmentation. 
% The resulting Vision Transformer, fine-tuned on this extensive dataset, will be released to facilitate future research in physically-grounded 3D content. 
For the task of dense material segmentation, we selected the Text-Image Pretraining with Spatial awareness (TIPS) model \cite{maninis2024tips}. This choice was motivated by TIPS's design, which combines contrastive image-text learning with self-supervised masked image modeling \cite{he2022masked, bao2021beit} to produce spatially coherent representations. This makes its features well-suited for dense prediction tasks like segmentation, where it has been shown to achieve state-of-the-art performance \cite{maninis2024tips}.

For our 2D segmentation module, we adopt a parameter-efficient fine-tuning approach. We freeze the publicly available pre-trained TIPS encoder and fine-tune TIPS DPT decoder \cite{ranftl2021vision} for dense segmentation on our curated material dataset \cite{maninis2024tips}. This approach benefits from the robust features learned during pre-training, while ensuring training efficiency. The decoder was trained for 40K iterations using the AdamW optimizer \cite{loshchilov2017decoupled}, a batch size equal to 32, a learning rate of 0.0001 and a weight decay of 0.0001.

\subsubsection{Multi-View Segmentation Inference}
At test time, given a new input mesh $M$, we render it from 11 diverse viewpoints. This multi-view approach is necessary for capturing comprehensive appearance information and for resolving ambiguities in later stages (\Cref{subsec:2d_3d}). We then apply our fine-tuned TIPS-based model to each rendered image $I_i$ to produce a 2D material segmentation map $S_i$. As individual 2D views often contain ambiguities from lighting, reflections, or occlusions, the robust aggregation mechanism described next is required.

\subsection{2D-to-3D Consistency Projection}
\label{subsec:2d_3d}
% This is the core technical contribution: we leverage 3D consistency to resolve the errors from the previous step. 
% - Rendering face buffer: for each voviewpoint, we render a unique color for each face of the mesh $M$. This creates a colour buffer to identify which face of is visible at each pixel.
% - Vote casting: We iterate through each pixel of each of the N views. For a given pixel in view i, we look up the predicted material label $m = S_i(u,v)$ and the visible face index $f = F_i(u,v)$. We then cast one vote for material $m$ to be assigned to face $f$. After processing all pixels in all views, each face $f$ has accumulated a set of votes from every viewpoint where it was visible.
% - For each face $f$, we assign the material label that received the most votes. This majority vote filters out single-view errors and converges on the most globally consistent material. Faces that were never visible in any view (e.g., the underside of an object) are assigned a special "unseen" label.
To resolve ambiguities present in individual 2D segmentations, we employ a robust per-face aggregation strategy based on majority voting. This mechanism effectively filters single-view errors and converges on the most globally consistent material (\Cref{fig:projection}).

For each of the 11 viewpoints, we render a face index buffer, where each pixel is colored with a unique ID corresponding to the visible face of the mesh $M$. This allows us to establish a direct mapping from any pixel in a 2D render to a specific face on the 3D model.

We then implement a majority voting scheme. We iterate through each pixel $(u, v)$ in each of the 11 rendered views. For a given pixel in view $i$, we retrieve two elements: the predicted material label $m = S_i(u,v)$ from the 2D material segmentation map, and the visible face index $f$ from the corresponding face index buffer. We then cast a vote for material $m$ to be assigned to face $f$. After processing all pixels across all views, each face on the mesh is assigned with material votes from every viewpoint in which it was visible.

The final label $L_f$ for face $f$ is determined by 
\begin{equation}
L_f = \text{arg max}_{m \in \mathcal{M}} \sum_{i=1}^N [S_i(Proj(f, i)) = m] , 
\end{equation}
where $Proj(f, i)$ returns the pixel coordinates of the centroid of face $f$ in view $i$, and $\mathcal{M}$ is the set of material classes. While a similar multi-view majority voting concept is explored by Phys4DGen \cite{lin2024phys4dgen} for aggregating SAM2 geometric part masks over 3D Gaussians, our formulation differs in its semantic target. We project dense material identity predictions directly onto explicit mesh faces rather than tracking geometric masks across Gaussians. This ensures that disjoint geometric parts sharing the same material (e.g., four separated wooden chair legs) are consistently unified into the same material class.

This aggregation is highly effective in resolving ambiguities by filtering out single-view errors and converging on the most globally consistent material. A flood fill algorithm is used for faces that were never visible in any of the rendered views, assigning them a label equal to the majority neighboring faces.

\subsection{3D Contextual Refinement with a Vision-Language Model}
While our majority vote approach enforces geometric consistency, labels may still contain semantic errors due to out-of-distribution assets or lighting. To address these inconsistencies, we employ a Vision-Language Model (VLM) to critique and repair our initial 3D material assignments.

To achieve this, we adopt Chain-of-Thought prompting \cite{wei2022chain} with three inputs: (1) the original multi-view RGB images $\{I_i\}$, (2) our 2D segmentation masks after being re-projected into color-coded maps $\{I'_i\}$, and (3) an automatically generated legend mapping colors to material labels. The VLM (Gemini-2.5 Pro) is instructed to identify and correct material assignments that are semantically implausible given the object's identity and context. For instance, the VLM might recognize that the thin legs of a particular chair are more likely to be \textit{metal} than \textit{wood}, even if specular highlights caused the segmentation model to misclassify them. This step corrects semantic errors and ensures the final material assignments are both geometrically consistent and coherent with the object's identity and function. This approach aligns with recent findings that VLMs can self-correct semantic grounding errors when prompted with visual verification tasks \cite{liao2025can}. 

\subsection{Part Completion and Property Assignment}
% - Once every face in M has a material label, we group all connected faces with the same label to form surface patches. This results in a set of incomplete, non-watertight part meshes ${M'_1, ..., M'_k}$.
% - Amodal part completion: these parts must be converted to watertight meshes for physics simulation. We leverage HoloPart to convert each incomplete part $M'_i$ into a watertight mesh $P_i$. this is necessary to estimate volumetric properties.
% - We use a material lookup table to assign physical properties. For each material label (e.g., "wood"), we assign corresponding values for:
% Density, Restitution, Friction coefficients.
% The final output is a set of simulation-ready 3D objects that are both visually faithful to the generative model's output and physically plausible.
With a consistent and semantically valid material label assigned to every visible face, we proceed to generate the final simulation-ready components. First, we partition the mesh by grouping all connected faces with the same material label, resulting in a set of non-watertight surface patches $\{M'_1, ..., M'_k\}$. For accurate physics simulation, these surface patches must be converted into watertight volumes. To this end, we employ HoloPart \cite{yang2025holopart}, a state-of-the-art amodal completion model, to convert each incomplete shell $M'_i$ into a complete mesh $P_i$. This is essential for estimating volumetric properties like mass and for ensuring stable interactions in a physics engine. 

Once we have a set of watertight material-aware components, we assign them physical properties using a VLM (Gemini-2.5 Pro) to reason about the object's context. For each completed part, we prompt the VLM with multi-view renderings to infer two attributes: 
\begin{enumerate}[topsep=0pt]
    \item A plausible density ($\rho_i$), considering its material label and the overall object context. Optionally, we infer simulator-specific properties, such as the parameters \textit{solimp} and \textit{solref} for MuJoCo-based simulations \cite{todorov2012mujoco}.
    \item Its likely internal structure (\textit{Solid} or \textit{Hollow}). If hollow, the VLM also predicts an average wall thickness ($t_i$).
\end{enumerate}

\subsubsection{Inferring Internal Structure for Mass Calculation.}
The ability to differentiate between solid and hollow components is a key advantage over prior work. Lacking a complete mesh representation, volumetric approaches like NeRF2Physics assume a uniform thickness heuristic. This assumption is physically inaccurate for solid objects. In contrast, we employ the same thickness-based approximation only for parts predicted to be hollow, and perform a volume-based calculation for parts identified as solid.

The mass $m_i$ for each part is then calculated based on its predicted structure. For solid parts, we compute the mass directly from the mesh volume $V_i$ as $m_i = \rho_i \cdot V_i$. For hollow parts, where the internal geometry is unknown, we approximate the mass using the surface area $A_i$ and predicted thickness $t_i$ as $m_i \approx \rho_i \cdot A_i \cdot t_i$. The total mass of the object is the sum of its component masses $M_{total} = \sum_{i} m_i$. 

From these physically-grounded parts, other dynamic properties like the inertia tensor can be derived. The final output of Gen2Physics is a collection of simulation-ready 3D components that are both visually faithful to the generative model's output and physically plausible for interactive applications.
\section{Results}

We evaluate Gen2Physics on two tasks: (1) downstream physical property estimation and rigid-body simulation readiness, and (2) material segmentation fidelity.
\subsection{Datasets}
\begin{figure}[t!]
\centering{\includegraphics[width=0.4\textwidth]{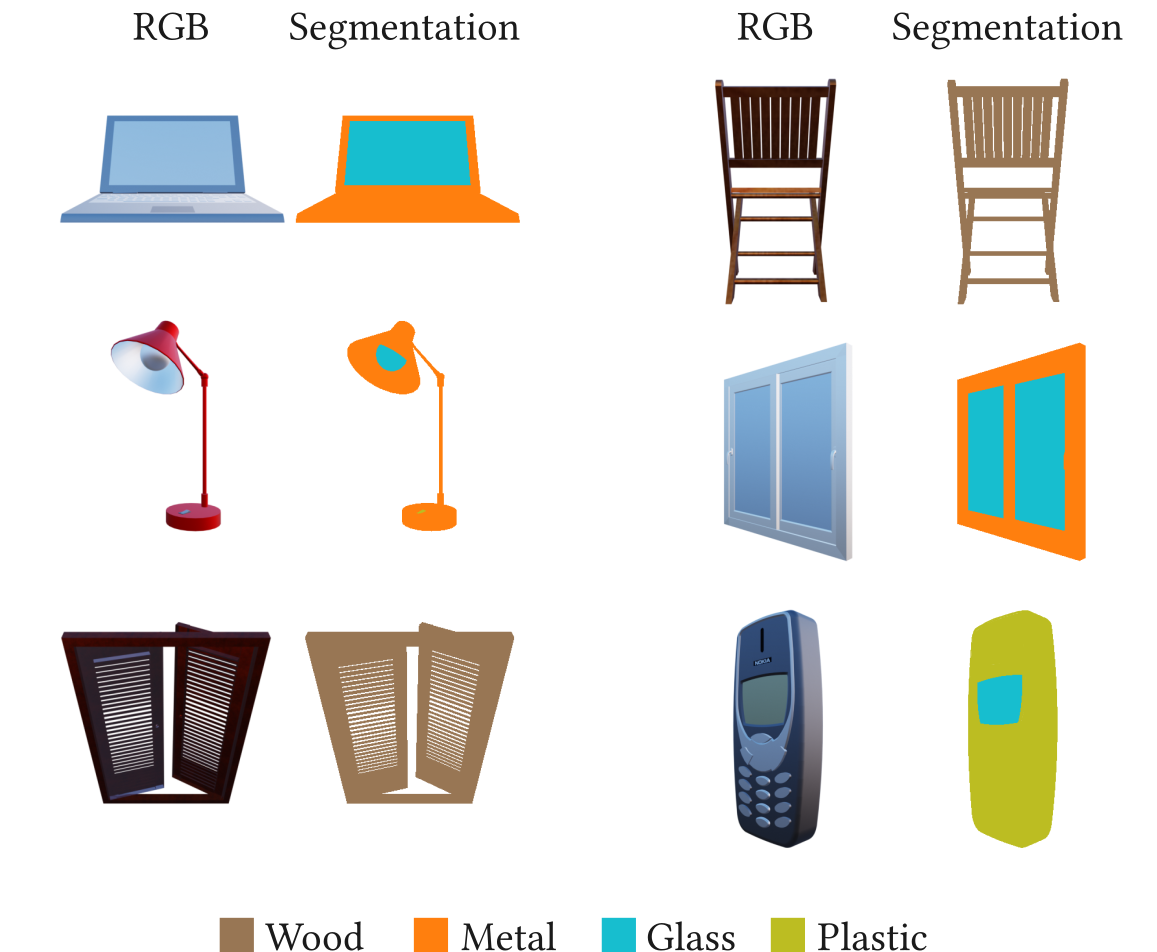}}
  \caption{Samples of the PartNet-Material dataset. Dense segmentation masks are obtained by manually assigning material labels to each individual part of the asset.}
  \label{fig:partnet}
\end{figure}

We use a combination of a standard benchmark for the downstream task of mass estimation, and a novel manually annotated dataset for segmentation.
\paragraph{ABO-500 Benchmark} To assess the utility of our framework for physics simulation, we evaluate mass estimation performance on the ABO-500 benchmark. This subset of the Amazon Berkeley Objects (ABO) \cite{collins2022abo} dataset contains 3D models of real-world products, complete with their ground-truth mass. All methods are evaluated on the same ground-truth ABO assets.
\paragraph{PartNet-Material Dataset} For the task of material segmentation, we introduce the PartNet-Material dataset. We rendered 11 different viewpoints from 100 multi-material PartNet objects \cite{mo2019partnet}, and annotated 1100 pixel-aligned segmentation masks. This dataset serves as our primary benchmark for evaluating segmentation accuracy. Examples of RGB images and corresponding material segmentations are shown in \Cref{fig:partnet}.

\begin{figure*}[t!]
\centering{\includegraphics[width=0.90\textwidth]{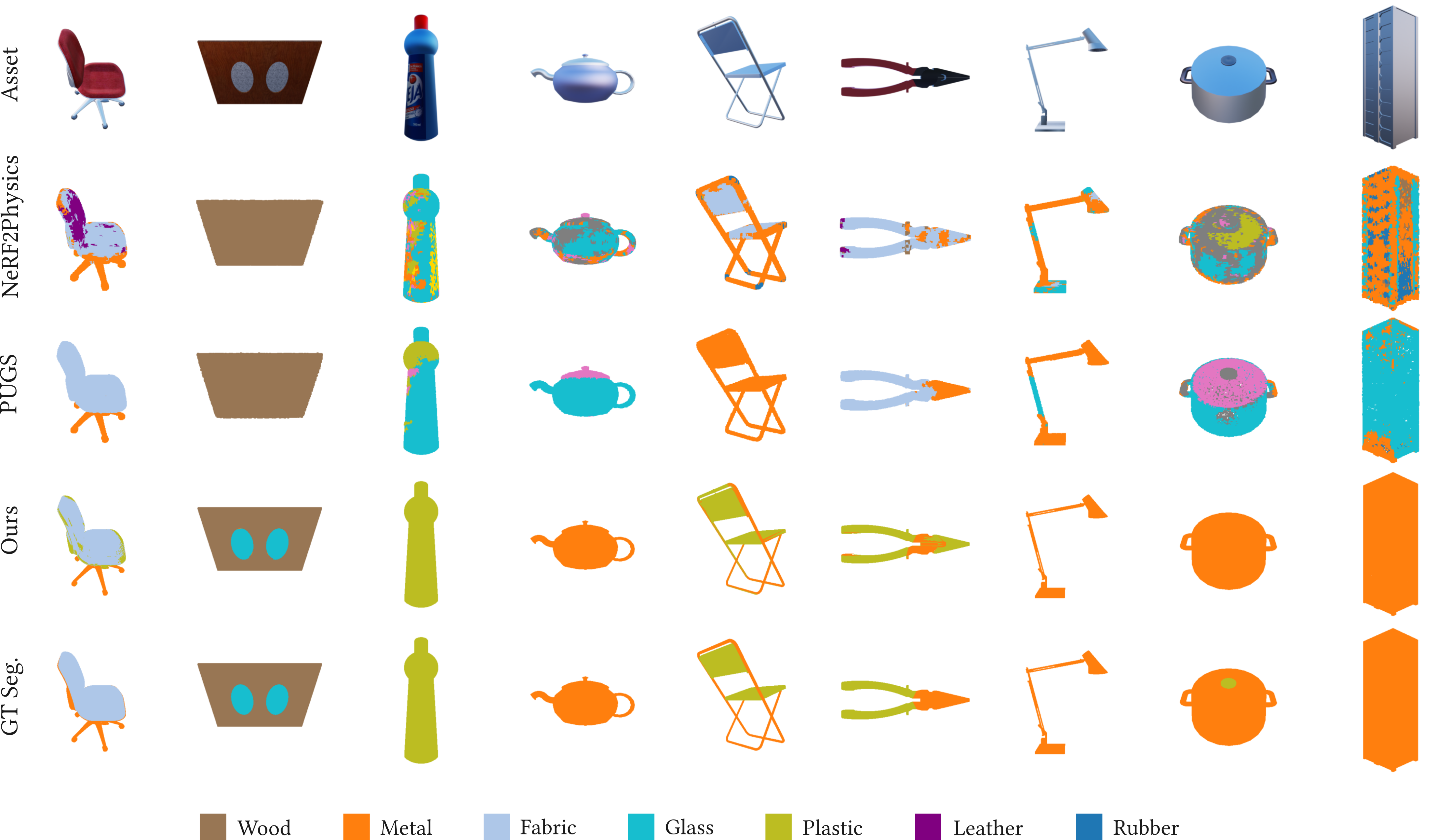}}
  \caption{We compare Gen2Physics (Ours), NeRF2Physics, and PUGS on a material identification task. The visualizations show that the 2D-to-3D consistency projection and VLM refinement proposed in Gen2Physics ensure more coherent and accurate material segmentations. In contrast, NeRF2Physics and PUGS produce fragmented masks due to their reliance on local CLIP embeddings}
  \label{fig:segmentation}
\end{figure*}

\subsection{Mass estimation}

We evaluate Gen2Physics in physics-aware tasks by comparing its mass estimation capabilities with state-of-the-art methods. Following NeRF2Physics~\cite{zhai2024physical} and PUGS~\cite{shuai2025pugs}, we evaluate our method on the ABO-500 dataset\cite{collins2022abo} and report on four standard metrics: Absolute Difference Error (ADE), Absolute Log Difference Error (ALDE), Absolute Percentage Error (APE), and Min Ratio Error (MnRE). As noted in prior work~\cite{zhai2024physical}, MnRE is particularly informative as it is robust to biases on heavy objects. 
% Focusing on outperforming the baselines
As reported in \Cref{tab:mass_prediction}, Gen2Physics achieves comparable results with the current state-of-the-art models within 0.2\% of ADE and placing second on ALDE, while simultaneously producing decomposed simulation-ready assets. Moreover, our method surpasses previous methods on the MnRE metric and confirms reliable prediction capability across the diverse objects in the ABO-500 dataset. While PUGS records a better APE score, prior work \cite{zhai2024physical} highlights that this metric is known to be sensitive to errors on lightweight objects, where small absolute deviations result in high errors. 

Overall, Gen2Physics material segmentation process, refined with VLM contextual reasoning, provides a more accurate basis for assigning material densities. Moreover, our mesh-based approach enables a more principled volume calculation than the thickness estimation heuristics required by volumetric methods, although it overestimates scores for very small objects.

\begin{table}[h!]
    \centering
    \small
    \setlength{\tabcolsep}{0.4pt}
    \caption{Mass estimation and asset usability on the ABO-500 benchmark. Gen2Physics obtains the best MnRE, and is within 0.2\% of the best ADE, while being the only method to output watertight per-material sub-meshes (Sim. Ready) compatible with rigid-body physics engines. Baselines reported from \cite{shuai2025pugs, dagli2025vomp}.}
    \begin{tabular}{lccccc}
        \toprule
        Method & \shortstack{Sim.\\Ready} & ADE ($\downarrow$) & ALDE ($\downarrow$) & APE ($\downarrow$) & MnRE ($\uparrow$)\\
        \midrule
        NeRF2Physics & \xmark & 12.725 & 0.736 & 1.040 & 0.564 \\
        PUGS & \xmark & 9.461 & 0.661 & \first{0.767} & \second{0.576} \\
        VoMP & \xmark & \first{8.433} & \first{0.631} & 0.887 & \second{0.576} \\
        Phys4DGen & \xmark & 9.961 & 0.664 & \second{0.825} & 0.566 \\
        \textbf{Gen2Physics (Ours)} & \cmark & \second{8.447} & \second{0.651} & 1.170 & \first{0.600} \\
        \bottomrule
    \end{tabular}
    \label{tab:mass_prediction}
\end{table}

\subsection{Material identification}

\begin{table*}[t!]
    \centering
    \caption{Per-class Intersection-over-Union (IoU [\%] $\uparrow$) and mean IoU (mIoU [\%] $\uparrow$) on the PartNet-Material benchmark. The 8 reported categories are the foreground material classes populated across the test split. Organic and Leather never occur and are excluded, while Paper has 0.00 IoU for all methods and is omitted for space. Overall mIoU is calculated as the unweighted mean across these active classes. Colors denote the \colorbox{first_color}{best} and \colorbox{second_color}{second-best} results respectively.}
    \begin{tabular}{lccccccc|c}
        \toprule
        Method & Wood & Fabric & Metal & Plastic & Glass & Ceramic & Stone & mIoU [\%] $\uparrow$ \\
        \midrule
        NeRF2Physics & 61.87 & 2.67 & 23.23 & 1.03 & \second{6.12} & 0.00 & 21.43 & 14.54 \\
        PUGS & \second{69.50} & 4.94 & 15.66 & 0.61 & 3.90 & 0.00 & \first{29.84} & 15.56 \\
        \midrule
        Gen2Physics (ViT only) & 66.50 & \second{10.88} & \second{38.71} & \second{33.97} & 5.34 & 0.00 & \second{25.23} & \second{22.58} \\
        Gen2Physics (Full) & \first{86.23} & \first{35.72} & \first{84.38} & \first{71.93} & \first{40.11} & \first{43.49} & 24.32 & \first{48.27} \\
        \bottomrule
    \end{tabular}
    \label{tab:segmentation_results}
\end{table*}

To evaluate the task of material identification, we compare Gen2Physics, which relies on a fine-tuned ViT and a VLM reasoning step, against existing baselines that leverage CLIP embeddings, namely NeRF2Physics and PUGS. The comparison, performed on the manually annotated PartNet-Material dataset (\Cref{fig:segmentation}), is intended to demonstrate the performance ceiling of zero-shot vision-language embeddings versus dedicated dense material representations when lifted into 3D geometry.

To ensure a fair and controlled comparison, we isolate the quality of each method's material feature representation from confounding factors such as 3D reconstruction quality or LLM reasoning bias. We establish a standardized benchmark protocol across three dimensions. First, all methods are evaluated on the exact same ground-truth 3D mesh geometry and camera viewpoints. For NeRF2Physics, we compute CLIP features \cite{radford2021learning} directly on a point cloud of 30,000 points sampled uniformly across the ground-truth mesh surface. For PUGS, 3D Gaussians are initialized at these identical point locations before running its full optimization pipeline with its region-aware feature contrastive loss. Second, while baseline methods originally produce open-vocabulary embeddings, we constrain all methods to select from the same discrete material vocabulary (\Cref{subsubsec:data-collection}). Finally, we use the identical Vision-Language Model (Gemini 2.5 Pro \cite{comanici2025gemini}) with deterministic decoding (temperature $T=0.0$) across all methods to assign final material labels. This setup ensures that differences in mIoU reflect the discriminative power of the underlying visual representations (zero-shot CLIP embeddings vs. our fine-tuned dense ViT) rather than differences in geometry, rendering, or language decoders.

\begin{figure*}[t!]
\centering{\includegraphics[width=0.88\textwidth]{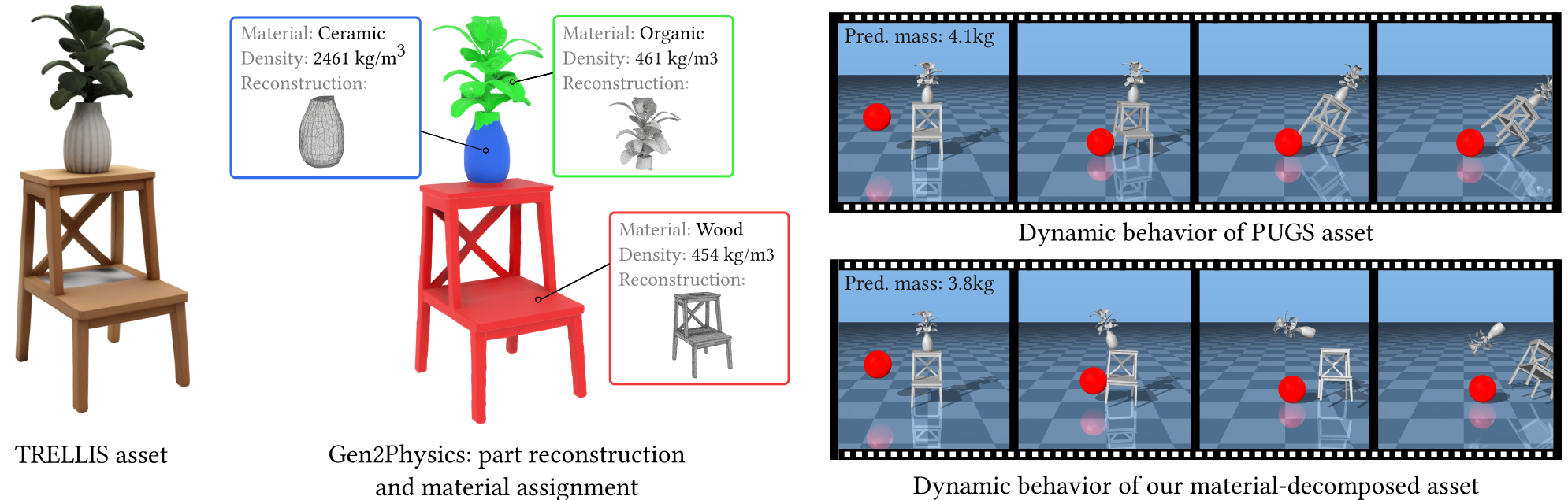}}
  \caption{Qualitative result showing part-reconstruction, material and physics assignment for a TRELLIS generated multi-composite asset. While both methods estimate plausible total mass, PUGS treats the mesh as a monolithic asset. In contrast, our framework decomposes and reconstructs watertight sub-meshes for a physically plausible dynamic behavior, as shown by the MuJoCo simulation \cite{todorov2012mujoco}.}
  \label{fig:dynamic}
\end{figure*}

\paragraph{Quantitative results} We measure performance on material identification by calculating the mean Intersection-over-Union (mIoU) across segmentation maps, following the evaluation performed in prior work \cite{cordts2016cityscapes, mmseg2020}. Specifically, we first aggregate a confusion matrix over the entire test dataset. From this matrix, we compute the Intersection-over-Union for each class $c$. The mIoU is calculated as the average of the per-class IoU scores.

\Cref{tab:segmentation_results} shows that Gen2Physics significantly outperforms baselines, achieving 48.27\% mIoU, which more than doubles the performance of PUGS (15.56\%) and NeRF2Physics (14.54\%). This improvement is consistent across nearly all categories. This confirms that using a Vision Transformer specifically fine-tuned for dense material segmentation provides a much stronger basis for material segmentation than general-purpose CLIP embeddings used by prior work. PUGS's region-aware loss improves coherence over NeRF2Physics, but its features are still based on CLIP computed on local patches, which do not effectively differentiate materials with similar textures or lighting.

\paragraph{Qualitative results}
\Cref{fig:segmentation} confirms the findings mentioned in the quantitative analysis. The baseline methods produce fragmented and noisy masks as direct result of comparing general-purpose CLIP features, extracted from local image patches, against the language embeddings of material names. This approach is suboptimal for two reasons: firstly, patch-level visual features (e.g. a specular highlight) do not always align with high-level semantic concepts (e.g. \textit{Wood}), leading to misclassifications. Secondly, local embeddings lack the broader context of the object, constraining the ability to ground the segmentation process on global understanding of the object. Our framework explicitly overcomes these limitations through its specialized ViT and subsequent VLM refinement.

\subsection{Ablations}
\begin{table}[t]
 \centering
    \caption{Ablation study on the ABO-500 mass estimation benchmark. Results show that internal structure reasoning is important for achieving robust mass calculation.}
 \resizebox{1 \columnwidth}{!}{% 
        \setlength{\tabcolsep}{1.4pt}
    \begin{tabular}{ccccc}
    \toprule
    Method & ADE ($\downarrow$) & ALDE ($\downarrow$) & APE ($\downarrow$) & MnRE ($\uparrow$)\\
    \midrule
    w/o Solid/Hollow & \first{7.636} & 0.699  & 2.189 & 0.595 \\
    Full & 8.447 & \first{0.651} & \first{1.170} & \first{0.600} \\
    \bottomrule
    \end{tabular}
    }
    \label{tab:ablations}
\end{table}

To isolate the contributions of our key components, we conducted a series of ablations (\Cref{tab:segmentation_results}, \Cref{tab:ablations}). First, we remove the VLM-based contextual refinement step, as shown in \Cref{tab:segmentation_results} (\textit{Gen2Physics (ViT only)}). Even without VLM refinement, our method achieves an mIoU of 22.58\%, outperforming both baselines. The full Gen2Physics model, which re-introduces VLM refinement, shows an increase in mIoU to 48.27\%. This shows that, while our ViT acts as a powerful feature extractor, the VLM's contextual reasoning is critical for correcting semantic errors, especially on out-of-distribution assets or visually-ambiguous materials. The refinement also recovers classes the dense predictor misses outright: Ceramic is never correctly segmented by the ViT alone (0.00 IoU) but reaches 43.49 IoU after refinement, indicating that object-level context resolves materials whose local appearance is ambiguous against Plastic and Stone.

Next, we assess the impact of reasoning about an object's internal structure. In the \textit{Ours w/o Solid/Hollow} ablation (\Cref{tab:ablations}), we disabled the VLM prediction and assumed all volumes to be solid. Interestingly, assuming all parts are solid improves the ADE score, but causes performance to degrade on the ALDE, APE, and MnRE metrics. This suggests that, while our solid/hollow reasoning provides a more principled physical model, a naive "all-solid" heuristic performs better on the several simple heavy objects in the ABO dataset. However, our full model's performance on the relative ALDE, APE, and MnRE metrics demonstrates its generality and robustness. We believe improving the VLM's internal structure prediction is a promising avenue for future work (\Cref{sec:conclusion}).

\subsection{Qualitative physics simulation}
A core downstream task enabled by our framework is plausible dynamic simulations from generated assets. \Cref{fig:dynamic} demonstrates the difference in behavior for an asset processed by our method versus the strongest baseline (PUGS). While both methods estimate comparable total mass ($m_{PUGS}=4.1kg$, $m_{Ours}=3.8kg)$, PUGS treats the asset as a monolithic object. In contrast, Gen2Physics decomposes the asset into its constituent parts, leading to a dynamic behavior that matches real-world intuition. More information can be found in the videos provided in the Supplementary material.

\section{Limitations and Conclusions}
\label{sec:conclusion}
In this work we introduced Gen2Physics, a novel framework that combines the visual fidelity of modern 3D generative models with the physical plausibility required for interactive applications. Our proposed 2D-to-3D consistency projection mechanism aggregates multi-view segmentations into a coherent material decomposition of a 3D mesh, while a subsequent VLM refinement step ensures semantic validity. By completing these material-aware components into watertight volumes and reasoning about their internal structure, Gen2Physics produces simulation-ready assets with accurate physical properties. 

Gen2Physics currently has limitations that suggest avenues for future research. Our framework currently relies on the external model HoloPart for amodal part completion. Future work should integrate this step into a single end-to-end pipeline. Moreover, similar to limitations highlighted in prior works \cite{lin2024phys4dgen}, our current method focuses primarily on surface-visible material decomposition. Future work could explore mechanisms for inferring and modelling internal material structures, which are critical for objects with heterogeneous compositions. We also note that our manually-annotated PartNet-Material test set, while precise, is limited to 1100 images. The creation of a larger benchmark remains an important direction for future work. Despite these limitations, Gen2Physics represents a step towards creating a \textit{seamless} pipeline from generative AI to interactive, physically realistic 3D content.

{
    \small
    \bibliographystyle{ieeenat_fullname}
    \bibliography{main}
}


\end{document}